\documentclass{article}

\usepackage{microtype}
\usepackage{graphicx}
\usepackage{subfigure}
\usepackage{booktabs} 
\usepackage{amsmath}

\usepackage{hyperref}

\usepackage[accepted]{icml2020}
\usepackage{caption}
\usepackage[frozencache=true,cachedir=.]{minted}

\icmltitlerunning{Improving Few-Shot Learning with Auxiliary Self-Supervised Pretext Tasks}

\begin{document}

\twocolumn[
\icmltitle{
    Improving Few-Shot Learning with\\
    Auxiliary Self-Supervised Pretext Tasks
}



\icmlsetsymbol{equal}{*}

\begin{icmlauthorlist}
\icmlauthor{Nathaniel Simard}{equal,udem,mila}
\icmlauthor{Guillaume Lagrange}{equal,udem,mila}
\end{icmlauthorlist}

\icmlaffiliation{udem}{Department of Computer Science and Operations Research, University of Montreal, Montreal, QC, Canada}
\icmlaffiliation{mila}{Mila – Quebec Artificial Intelligence Institute, Montreal, QC, Montreal, Canada}

\icmlcorrespondingauthor{Nathaniel Simard}{nathaniel.simard@umontreal.ca}
\icmlcorrespondingauthor{Guillaume Lagrange}{guillaume.lagrange@umontreal.ca}

\icmlkeywords{Machine Learning, ICML}

\vskip 0.3in
]



\printAffiliationsAndNotice{\icmlEqualContribution} 

\begin{abstract}
Recent work on few-shot learning \cite{tian2020rethinking} showed that quality of learned representations plays an important role in few-shot classification performance. On the other hand, the goal of self-supervised learning is to recover useful semantic information of the data without the use of class labels. In this work, we exploit the complementarity of both paradigms via a multi-task framework where we leverage recent self-supervised methods as auxiliary tasks. We found that combining multiple tasks is often beneficial, and that solving them simultaneously can be done efficiently. Our results suggest that self-supervised auxiliary tasks are effective data-dependent regularizers for representation learning. Our code is available at: \url{https://github.com/nathanielsimard/improving-fs-ssl}.
\end{abstract}


\section{Introduction}
Few-shot learning measures the ability to learn new concepts from a limited amount of examples. This is a challenging problem that usually requires different approaches from the success stories of image classification where labeled data is abundant. Recently, meta-learning \cite{schmidhuber1987srl, bengio1992slr, thrun1998ltl}, or learning-to-learn, has emerged as a learning paradigm where a machine learning model gains experience over multiple learning episodes and uses this experience to improve its future learning performance. Significant advances in few-shot learning \cite{vinyals2016matching, ravi2016optimization, snell2017prototypical, finn2017model, oreshkin2018tadam, sung2018learning, rusu2019meta, lee2019meta} have been made by framing the problem within a meta-learning setting.


More precisely, few-shot learning can be seen as a sub-task of meta-learning where a learner is trained and tested on several different but still related tasks. The goal of few-shot meta-learning is to train a model that can quickly adapt to a new task using only a few datapoints and training iterations \cite{finn2017model}. During the meta-training phase, the model is trained to solve multiple tasks such that it is able to learn or adapt to new ones through the meta-testing phase. The performance of the learner is evaluated by the average test accuracy across many meta-testing tasks.


Focusing on few-shot image classification, where very few images are available for novel tasks, recent works aim to learn representations that generalize well to novel classes by training a feature representation to classify a training dataset of base classes \cite{finn2017model, vinyals2016matching, snell2017prototypical, gidaris2018dynamic, qi2018low}. Methods to tackle this problem include optimization-based methods and metric-based methods, although recent works suggest that learning a good representation is the main factor responsible for fast adaptation in few-shot learning methods \cite{raghu2020rapid, tian2020rethinking}. \citeauthor{raghu2020rapid} (\citeyear{raghu2020rapid}) empirically suggested that the effeciveness of MAML \cite{finn2017model} is due to its ability to learn a useful representation, while \citeauthor{tian2020rethinking} (\citeyear{tian2020rethinking}) showed that learning a good representation of the data through a proxy task is even more effective than complex meta-learning algorithms.

Self-supervised learning is another recent paradigm but for unsupervised learning where the supervisory signal for feature learning is automatically generated from the data itself. Seminal works \cite{doersch2015unsupervised, zhang2016colorful, noroozi2016unsupervised, gidaris2018unsupervised} have relied on heuristics to design \textit{pretext} learning tasks such that high-level image understanding must be captured to solve them. Discriminative approaches based on contrastive learning in the latent space \cite{oord2018representation, wu2018unsupervised, bachman2019learning, tian2019contrastive, henaff2020data, chen2020simple, he2020momentum, chen2020improved, misra2020self, chen2020big} have shown recently that self-supervised learning is especially useful with the large availability of unlabeled data, effectively closing the gap with supervised learning when leveraging models with large capacity. These methods are trained by reducing the distance between different augmented views of the same image (positive pairs), and increasing the distance between augmented views of different images (negative pairs). More recently, BYOL \cite{grill2020bootstrap} showed that one can also learn transferable visual representations via bootstrapping representations and without negative pairs.

In this paper, we propose to leverage self-supervised learning method(s) as auxiliary task(s) to prevent overfitting to the base classes seen during meta-training and improve transfer learning performance on novel classes. More specifically, we build upon the simple baseline proposed by \citeauthor{tian2020rethinking} (\citeyear{tian2020rethinking}), where the meta-training tasks are merged into a single pre-training task and a \textit{linear} model is learned on top of the frozen encoder, and leverage self-supervised auxiliary task(s) as a data-dependant regularizer to learn richer and more transferable visual representations. Our multi-task framework exploits the complementarity of few-shot learning and self-supervised learning paradigms to boost performance on few-shot image classification benchmarks.


 

\section{Related Work}
\paragraph{Few-shot learning.} Recent works have mostly focused on meta-learning approaches to few-shot learning. Among these, optimization-based methods \cite{finn2017model, ravi2016optimization, lee2019meta, rusu2019meta} aim to learn how to rapidly adapt the embedding model parameters to training examples for a given few-shot recognition task. Metric-based approaches \cite{vinyals2016matching, snell2017prototypical, sung2018learning, gidaris2018dynamic} aim to learn a task-dependent metric over the embeddings. More concretely, such methods learn a distance metric between a query image and a set of support images given a few-shot task.

More recently, it has been shown that learning a good representation plays an important role in few-shot learning \cite{raghu2020rapid, tian2020rethinking}. Our work is directly related to that of \citeauthor{tian2020rethinking}(\citeyear{tian2020rethinking}). A simple image classification task is used to train on the merged meta-training data to learn an embedding model, which is re-used at meta-testing time to extract embedding for a linear classifier.

\paragraph{Self-supervised learning.}
Deep learning models are usually considered data hungry and require a significant amount of labeled data to achieve decent performance. This recent learning paradigm aims to mitigate the requirements for large amounts of annotated data by providing surrogate supervisory signal from the data itself. Initial works in self-supervised representation learning relied on engineered prediction tasks \cite{doersch2015unsupervised, zhang2016colorful, noroozi2016unsupervised, gidaris2018unsupervised}, often referred to as pretext tasks. Such methods include predicting the relative position of image patches \cite{doersch2015unsupervised, noroozi2016unsupervised}, the rotation degree applied to an image \cite{gidaris2018unsupervised}, the colors of a grayscale image \cite{zhang2016colorful, larsson2016learning}, and many others.

More recently, discriminative approaches based on contrastive learning have shown great promise in self-supervised learning. Contrastive methods like MoCo \cite{he2020momentum, chen2020improved} and SimCLR \cite{chen2020simple, chen2020big} push representations of different views of the same image (positive pairs) closer, and spread representations of views from different images (negative pairs) apart. Beyond contrastive learning, BYOL \cite{grill2020bootstrap} relies only on positive pairs. From a given representation, referred to as \textit{target}, an \textit{online} network is trained to learn an enhanced representation by predicting the target representation. Although concerns over stability and representational collapse have been raised \cite{fetterman2020understanding, tian2020understanding}, BYOL has still proven to be an effective self-supervised representation learning technique.

\paragraph{Multi-task learning.}
Our work is also related to multi-task learning, a class of techniques that train on multiple task objectives together to learn a representation that works well for every task, though we are mainly interested in improving performance on the main task in this work. In this context, using multiple self-supervised pretext tasks is especially attractive as no additional labels are required. The combination of complementary tasks has been shown to be beneficial \cite{doersch2017multi, yamaguchi2019multiple} in self-supervised representation learning. Previous works \cite{gidaris2019boosting, su2020does} explore the use of self-supervised pretext tasks as auxiliary tasks to improve metric-based few-shot learning methods. In this work, we extend these ideas to recent self-supervised techniques and focus on learning richer and more generalizable features by starting from the simpler few-shot learning baseline of \citeauthor{tian2020rethinking} (\citeyear{tian2020rethinking}).

\section{Method}

In this section, we first describe in \S\ref{sec:fs-problem} the few-shot learning problem addressed and introduce in \S\ref{sec:multi-task} the proposed multi-task approach to improve few-shot performance with self-supervised auxiliary tasks.

\begin{figure*}[t]
    \centering
    \includegraphics[width=0.8\textwidth]{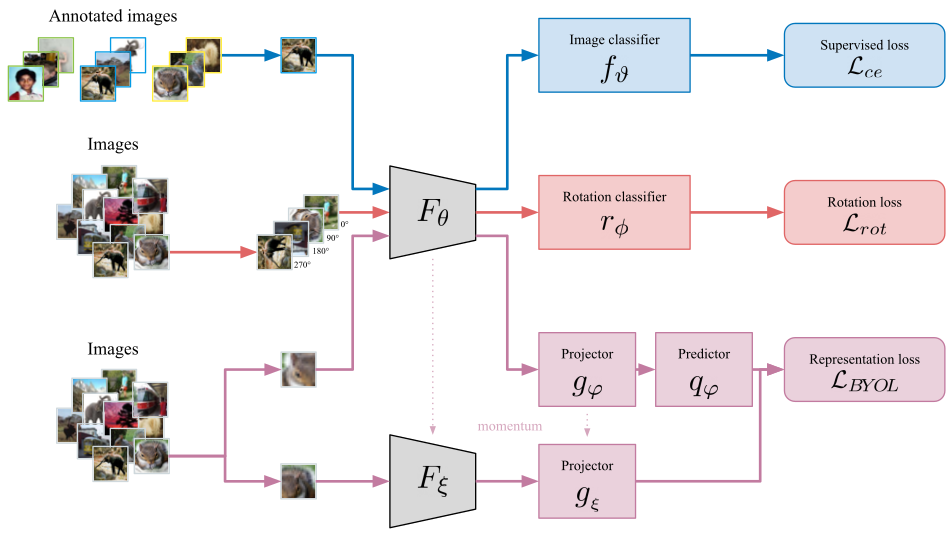}
    \caption{\textbf{Combining supervised and self-supervised tasks for meta-training.} We train the embedding model $F_\theta$ with both annotated images and unlabeled images in a multi-task setting. Self-supervised tasks such as rotation prediction or representation prediction (BYOL) act as a data-dependent regularizer for the shared feature extractor $F_\theta$. Although additional unlabeled data can be used for the self-supervised tasks, in this work we sample images from the annotated set.}
    \label{fig:our-framework}
\end{figure*}

\subsection{Problem Formulation} \label{sec:fs-problem}


Standard few-shot learning benchmarks evaluate models in episodes of $N$-way, $K$-shot classification tasks. Each task consists of a small number of $N$ classes with $K$ training examples per class. Meta-learning approaches for few-shot learning aim to minimize the generalization error across a distribution of tasks sampled from a task distribution. This can be thought of as learning over a collection of tasks $\mathcal{T} = \{\left(\mathcal{D}_{i}^{train}, \mathcal{D}_{i}^{test}\right)\}_{i=1}^{I}$, commonly referred to as the \textit{meta-training} set.

In practice, a task is constructed on the fly during the meta-training phase and sampled as follows. For each task, $N$ classes from the set of training classes are first sampled (with replacement), from which the training (\textit{support}) set $\mathcal{D}_{i}^{train}$ of $K$ images per class is sampled, and finally the test (\textit{query}) set $\mathcal{D}_{i}^{test}$ consisting of $Q$ images per class is sampled. The \textit{support} set is used to learn how to solve this specific task, and the additional examples from the \textit{query} set are used to evaluate the performance for this task. Once the meta-training phase of a model is finished, its performance is evaluated on a set of held-out tasks $\mathcal{S} = \{\left(\mathcal{D}_{j}^{train}, \mathcal{D}_{j}^{test}\right)\}_{j=1}^{J}$, called the \textit{meta-test} set. During meta-training, an additional held-out \textit{meta-validation} set can be used for hyperparameter selection and model selection. Training examples $\mathcal{D}^{train} = \{\left(\mathbf{x}_t, y_t\right)\}_{t=1}^{T}$ and testing examples $\mathcal{D}^{test} = \{\left(\mathbf{x}_q, y_q\right)\}_{q=1}^{Q}$ are sampled from the same distribution, and are mapped to a feature space using an embedding model $F_{\theta}$. A base learner is trained on $\mathcal{D}^{train}$ and used as a predictor on $\mathcal{D}^{test}$.

\paragraph{Merged meta-training.} Following the simple baseline of \citeauthor{tian2020rethinking} (\citeyear{tian2020rethinking}), we merge tasks from the meta-training set into a single bigger task or dataset

\begin{equation}
\begin{split}
    \mathcal{D}^{merge} &= \{\left(\mathbf{x}_i, y_i\right)\}_{k=1}^{K}\\
    &= \cup\{\mathcal{D}^{train}_{1}, \ldots, \mathcal{D}^{train}_{i}, \ldots, \mathcal{D}^{train}_I\} \quad,
\end{split}
\end{equation}
where $\mathcal{D}^{train}_i$ is a task from $\mathcal{T}$.

The objective is to learn a transferable embedding model $F_\theta$ which generalizes to new tasks. For a task $\left(\mathcal{D}_{j}^{train}, \mathcal{D}_{j}^{test}\right)$ sampled from the meta-testing distribution, we freeze the embedding model $F_{\theta}$ to extract embeddings and train a base learner on $\mathcal{D}_{j}^{train}$. The base learner is instantiated as a simple linear classifier and is re-initialized for every task.

\subsection{Multi-Task Learning} \label{sec:multi-task}
Through solving non-trivial self-supervised tasks, the embedding model is encouraged to learn rich and generic image features that can be readily exploited for few-shot learning on novel classes. We propose a multi-task learning approach to extend the supervised objective with different self-supervised auxiliary tasks, where the goal is to learn a representation that works well for every task, and perhaps share knowledge between tasks, to further improve few-shot performance.

We incorporate self-supervision to our multi-task framework by adding auxiliary losses for each self-supervised task. As illustrated in Figure \ref{fig:our-framework}, we use the same embedding model $F_\theta$ to extract image features for all tasks. Based on the features extracted, a new network defined for each task is assigned to solve its respective task, each of which contribute to the overall loss

\begin{equation}
    \mathcal{L}_{tot} = \sum_{t=1}^N \mathcal{L}_t \quad,
\end{equation}
where $\mathcal{L}_t$ is the loss for the $t$-th task out of all $N$ tasks.

Along with the supervised objective, we consider two additional tasks in the present work: rotation prediction \cite{gidaris2018unsupervised} and representation prediction from different views following BYOL \cite{grill2020bootstrap}. All tasks are computed on the merged dataset $\mathcal{D}^{merge}$.

\paragraph{Supervised.} The supervised task is the standard classification task on all categories from the merged dataset, which has been shown to be an effective baseline to generate powerful embeddings for the downstream base learner \cite{tian2020rethinking}. The new network $f_\vartheta$ introduced to solve this task is instantiated as a simple linear classifier. We compute the cross-entropy loss $\mathcal{L}_{ce}$ between predictions and ground-truth labels.

\paragraph{Rotation.} In this task, the network must identify the rotation transformation applied to an image among four possible 2D rotations $\mathcal{R} = \{0^{\circ}, 90^{\circ}, 180^{\circ}, 270^{\circ}\}$, as proposed by \citeauthor{gidaris2018unsupervised} (\citeyear{gidaris2018unsupervised}). This auxiliary task is similar to \citeauthor{gidaris2019boosting} (\citeyear{gidaris2019boosting}), except that we do not create four rotated copies of an input image, which effectively quadruples the batch size, but instead sample a single rotation to apply for each image in the batch. The network $r_\phi$ specific to the rotation task is a multi-layer perceptron (MLP). The self-supervised loss $\mathcal{L}_{rot}$ for this task is the cross-entropy loss between the rotation predictions and the generated labels.

\paragraph{BYOL.} In BYOL \cite{grill2020bootstrap}, the \textit{online} network directly predicts the output of one view from another view given by the \textit{target} network. Essentially, this is a representation prediction task in the latent space, similar to contrastive learning except that it only relies on the \textit{positive} pairs. In this task, the online network is composed of the shared encoder $F_\theta$, the MLP projection head $g_\varphi$ and the predictor $q_\varphi$ (also an MLP). The target network has the same architecture as the online network (minus the predictor), but its paramters are an exponential moving average (EMA) of the online network parameters as illustrated in Figure \ref{fig:our-framework}. Denoting the parameters of the online network as $\theta_o = \{\theta, \varphi\}$, those of the target network as $\xi$ and the target decay rate $\tau \in [0,1)$, the update rule for $\xi$ is:
\begin{equation}
    \xi \leftarrow \tau \xi + (1 - \tau)\theta_o
\end{equation}
The self-supervised loss $\mathcal{L}_{B\!Y\!O\!L}$ is the mean squared error between the normalized predictions and target projections as defined in \citeauthor{grill2020bootstrap} (\citeyear{grill2020bootstrap}). Effectively, this task enforces the representations for different views of positive pairs to be closer together in latent space, which provides transformation invariance to the pre-defined set of data augmentations used for BYOL.

Although the multi-task framework proposed is very flexible and could easily be extended to additional tasks, the naïve implementation which does not share the transformed inputs for each task still has some overhead with each task added. Effectively, for each set of transformations associated to the different tasks, different views of the same input image are generated, which is essentially a new input in the batch. For example, given an input image from the merged dataset, the default data augmentations from the supervised task would create an augmented view of the input used to compute the supervised loss, and applying a rotation to the input image would generate a different view of the input. Not only does this double the batch size, but the inputs are not actually shared across tasks to compute the different losses. If instead the set of transformations are shared across tasks, the different tasks would be solved for the same inputs. In this scenario, the different losses are computed and backpropagated on the same inputs, which is much more efficient. In \S\ref{sec:exp-data-aug}, we explore this more efficient setting by combining the supervised and BYOL tasks using the stronger data augmentation strategy from BYOL in both. More concretely, we generate an augmented view of an input image and compute the supervised loss on the first augmented view, while another augmented view of the same input is generated to solve the representation prediction task in BYOL.


\section{Experimental Results} \label{sec:experiments}
In this section, we evaluate our proposed multi-task framework on two widely used few-shot image recognition benchmarks: miniImageNet \cite{vinyals2016matching} and CIFAR-FS \cite{bertinetto2018metalearning}.

\paragraph{Datasets.} The miniImageNet dataset \cite{vinyals2016matching} has recently become a standard benchmark for few-shot learning algorithms. It consists of 100 classes randomly sampled from ImageNet, with each class containing 600 down-sampled images of size $84 \times 84$. The CIFAR-FS dataset \cite{bertinetto2018metalearning} is derived from the original CIFAR-100 dataset by randomly splitting 100 classes into 64 classes for training, 16 for validation and 20 for testing. Each image is $32 \times 32$ pixels. In our experiments, we up-sampled the images to $96 \times 96$ pixels\footnote{This is due to the limitations of using smaller scale images with BYOL, which strongly relies on random crops as image augmentation. Initial experimental results showed that BYOL on its own could only achieve $47.09 \pm 0.96\%$ accuracy on CIFAR-FS, which is significantly lower than the results reported in Table \ref{tab:cifar-fs-results}.}.

\paragraph{Evaluation metrics.} As mentioned in \S\ref{sec:fs-problem}, few-shot algorithms are evaluated on a large number of $N$-way $K$-shot classification tasks. Each task is created by randomly sampling $N$ novel classes from the meta-test set, and then within the selected images randomly selecting $K$ training (support) images and $Q$ test (query) images (without overlap). In this work, we focus on $N=5$, $K=5$ (5-way 5-shot) classification, using the remainder of the $Q$ test samples to evaluate performance on the sampled task.

\subsection{Implementation Details} \label{sec:implementation}

\paragraph{Network architectures.} We mainly conduct our experiments using a ResNet-18 \cite{he2016deep} as our embedding model $F_\theta$ due to resource limitations, with an output feature vector of size $512$. Following previous works \cite{tian2020rethinking, snell2017prototypical, lee2019meta, oreshkin2018tadam, dhillon2020baseline}, we also report some results using ResNet-12 as our backbone for comparison. Our ResNet-12 is identical to that used in \citeauthor{tian2020rethinking} (\citeyear{tian2020rethinking}). Note that this modified ResNet-12 is wider than the original ResNet-18 and makes use of DropBlock \cite{ghiasi2018dropblock} as a regularizer, which results in a substantial gain in performance over ResNet-18 but at the cost of longer training time. Additionally, the resulting embedding size is $640$.

\paragraph{Task-specific networks.} From these embeddings, the image classifier $f_\vartheta$ is instantiated as a simple linear layer for classification. The rotation network $r_\phi$ is an MLP which consists in a linear layer with output size the same as the embedding size, followed by batch normalization, rectified linear units (ReLU), a linear layer with output size the same as the embedding size, and a final linear layer with output dimension equal to the number of rotation degrees to classify (here, 4). For BYOL, we use the following configurations. The projection MLP $g_\varphi$ consists in a linear layer with output size $2048$ followed by batch normalization \cite{ioffe2015batch}, rectified linear units (ReLU) \cite{nair2010rectified}, and a final linear layer with output dimension $128$. The predictor $q_\varphi$ uses the same architecture as $g_\varphi$. The target encoder $F_\xi$ uses the same architecture as the embedding model $F_\theta$, and the target projection head $g_\xi$ uses the same architecture as $g_\varphi$.

\paragraph{Optimization setup.} As an optimizer, we adopt SGD with a momentum of $0.9$ and a weight decay of $5e^{-4}$. The initial learning rate is $0.05$ and decayed by a factor of $0.1$ according to a multi-step learning schedule. All models are trained for $90$ epochs on CIFAR-FS with a decay step at epochs $45$, $60$ and $75$, except when BYOL is used (alone or in combination with other tasks). In this setting, we only decay the learning rate at epochs $60$ and $80$ since BYOL usually converges more slowly. On miniImageNet, all models are trained for $100$ epochs with a decay step at epochs $60$ and $80$, regardless of task combinations. For BYOL, the exponential moving average parameter $\tau$ is set to $0.99$. We use a batch size of $128$ for all experiments.

\paragraph{Data augmentation.} During pre-training on the merged meta-training dataset, we adopt random crop, color jittering and random horizontal flip as in \citeauthor{tian2020rethinking} (\citeauthor{tian2020rethinking}). For stronger data augmentation in BYOL, we adopt random crop, random color jittering, random grayscale, random horizontal flip and random gaussian blur similar to \citeauthor{fetterman2020understanding} (\citeyear{fetterman2020understanding}). Given that each task may require its own set of transformations, we use \texttt{Kornia} \cite{riba2020kornia} for efficient GPU-based data augmentation in \texttt{PyTorch} \cite{pytorch2019neurips}. Additional details can be found in Appendix \ref{appendix:data-aug}.

\paragraph{Early stopping.} For training the base learner, we use the same optimizer setup with SGD, and train for a maximum of $300$ steps, stopping early if the loss reaches a small threshold on the support set. Model selection is done on the held-out meta-validation set.

\subsection{Self-Supervision Alone is not Enough}
We first evaluate self-supervised representation learning with the two tasks explored in this work, rotation prediction and representation prediction (BYOL). Our results on CIFAR-FS (Table \ref{tab:cifar-fs-results}) and miniImageNet (Table \ref{tab:mini-imagenet-results}) show that, on its own, the rotation prediction pretext task limits the generality of the learned representations, significantly lagging behind in few-shot accuracy. On the other hand, the representation prediction task from different views (BYOL) shows great promise in representation learning, achieving results that are not that far from the supervised baseline\footnote{Note that we did not perform any extensive experiments to search for the optimal hyperparameter and learning schedule configurations for BYOL. Thus, the gap between BYOL and the supervised baseline could most likely be closed even further, perhaps just by adopting a longer training schedule.}.

Note that the performance gap between BYOL and the supervised baseline on miniImageNet is bigger than the one observed on CIFAR-FS. We attribute this to the slow convergence of BYOL. As noted in \S\ref{sec:implementation}, we changed the learning schedule for CIFAR-FS but did not do so for miniImageNet. Thus, we expect the gap to be smaller on miniImageNet with slightly longer training.

\begin{table}[th!]
    \caption{\textbf{Evaluating representation learning methods for few-shot recognition on CIFAR-FS.} 
    Average 5-way 5-shot classification accuracies on the test set with $95\%$ confidence intervals. We evaluate our method with 2 runs, where in each run the accuracy is the mean accuracy of 250 randomly sampled tasks. Models with $^{\dagger}$ use stronger data augmentation for the supervised objective. \textit{Sup.} refers to the supervised task, \textit{Rot.} to the rotation prediction task and \textit{BYOL} to the representation prediction task in \citeauthor{grill2020bootstrap} (\citeyear{grill2020bootstrap}).}
    \centering
    \begin{tabular}{llc}
    \toprule
    Method & Backbone & Accuracy (\%) \\
    \midrule
    Sup.               & ResNet-18 & $79.54 \pm 0.92$\\
    Rot.               & ResNet-18 & $49.26 \pm 1.01$\\
    BYOL               & ResNet-18 & $70.50 \pm 0.99$\\
    Sup. + Rot.        & ResNet-18 & $79.96 \pm 0.93$\\
    Sup. + BYOL        & ResNet-18 & $81.41 \pm 0.89$\\
    Sup. + BYOL + Rot. & ResNet-18 & $\mathbf{81.68} \pm \mathbf{0.98}$\\
    \midrule
    Sup.$^{\dagger}$           & ResNet-18 & $81.87 \pm 0.92$\\
    Sup.$^{\dagger}$+ BYOL    & ResNet-18 & $\mathbf{82.44} \pm \mathbf{0.91}$\\
    \bottomrule
    \end{tabular}
    \label{tab:cifar-fs-results}
\end{table}

\subsection{Combining Supervised and Self-Supervised Tasks} \label{sec:exp-combined}
On CIFAR-FS, we explore different combinations of tasks for our multi-task framework of Figure \ref{fig:our-framework}. In Table \ref{tab:cifar-fs-results}, our results show that both the rotation prediction task and BYOL improve the supervised baseline, though the biggest boost in performance is observed with BYOL. Additionally, combining all three tasks is even more beneficial. This result is intuitive, as we expect the features from the rotation prediction task to be complementary since there is no rotation transformation in any of the data augmentation strategies. In Table \ref{tab:mini-imagenet-results}, we show that the combination of all three tasks also improves the supervised baseline on miniImageNet.

\subsection{Data Augmentation: Stronger is Better} \label{sec:exp-data-aug}
In order to ensure that the performance improvement from BYOL is not strictly due to the stronger data augmentation strategy used by the task, we conduct experiments using the same data augmentations for the supervised baseline. An additional experiment without data augmentation for the supervised task is presented in Appendix \ref{appendix:exp-no-aug}.

On both CIFAR-FS (Table \ref{tab:cifar-fs-results}) and miniImageNet (Table \ref{tab:mini-imagenet-results}), we find that stronger data augmentation improves the supervised baseline. This is in line with a lot of the recent work in strong data augmentation techniques \cite{devries2017improved, zhang2018mixup, yun2019cutmix, cubuk2019autoaugment, cubuk2020randaugment}. Effectively, data augmentation is an important regularization technique that has been shown to improve generalization. Furthermore, we show that in this setting the addition of BYOL as a self-supervised auxiliary task still boosts the performance. As mentioned in \S\ref{sec:multi-task}, when used in combination with BYOL, both tasks share the same transformed inputs as part of our multi-task framework. Additional experiments where we leverage both augmented views generated to compute both the supervised and BYOL losses can be found in Table \ref{tab:extra-sup-byol} (Appendix \ref{appendix:extra-sup-byol}).

\begin{table}[th!]
    \caption{\textbf{Evaluating representation learning methods for few-shot recognition on miniImageNet.} 
    Average 5-way 5-shot classification accuracies on the test set with $95\%$ confidence intervals. We evaluate our method with a single run, where the accuracy is the mean accuracy of 250 randomly sampled tasks. Models with $^{\dagger}$ use stronger data augmentation for the supervised objective. \textit{Sup.} refers to the supervised task, \textit{Rot.} to the rotation prediction task and \textit{BYOL} to the representation prediction task in \citeauthor{grill2020bootstrap} (\citeyear{grill2020bootstrap}).}
    \centering
    \begin{tabular}{llc}
    \toprule
    Method & Backbone & Accuracy (\%) \\
    \midrule
    Sup.               & ResNet-18 & $67.26 \pm 0.89$\\
    Rot.               & ResNet-18 & $39.39 \pm 0.83$\\
    BYOL               & ResNet-18 & $51.20 \pm 0.97$\\
    Sup. + BYOL + Rot. & ResNet-18 & $\mathbf{68.59} \pm \mathbf{0.92}$\\
    \midrule
    Sup.$^{\dagger}$   & ResNet-18 & $69.42 \pm 0.94$\\
    Sup.$^{\dagger}$ + BYOL    & ResNet-18 & $\mathbf{71.47} \pm \mathbf{0.90}$\\
    \bottomrule
    \end{tabular}
    \label{tab:mini-imagenet-results}
\end{table}

\subsection{Comparison with Prior Work}
For comparison with prior work, we also report some results in Table \ref{tab:results-prior-work} using ResNet-12 as our backbone. Our supervised baseline is similar to the simple baseline of \citeauthor{tian2020rethinking} (\citeyear{tian2020rethinking}), except that we do not use any additional tricks like feature normalization, augmenting the number of support images, and our base learner is not a logistic regression classifier with L-BFGS optimizer from \texttt{scikit-learn} \cite{sklearn_api}. Due to resource limitations, we do not employ the additional rotation prediction task in our multi-task framework, though we expect it would slightly improve the reported performance based on our experiments in \S\ref{sec:exp-combined}.

On CIFAR-FS, our multi-task framework outperforms previous works by at least $1.5\%$. Note that our simple supervised baseline, even without the tricks used in \citeauthor{tian2020rethinking} (\citeyear{tian2020rethinking}), still performs better than their best distilled model. We attribute this to the higher resolution input images, which we upsampled to $96 \times 96$. On miniImageNet however, our simple baseline without their tricks is comparable to the results reported in their ablation study. We show that our framework improves this slightly weaker baseline, and anticipate that additional tricks such as feature normalization would complement our work to further improve the results.

Although \citeauthor{gidaris2019boosting} (\citeyear{gidaris2019boosting}) use a much larger network (WRN-28-10), we still include their results in Table \ref{tab:results-prior-work} for comparison as their work uses self-supervised rotation prediction as an auxiliary loss, which is very much related to our work. We expect a larger network to be more effective in learning richer features, and the results to improve with a larger architecture.

\begin{table*}[t]
    \caption{\textbf{Comparison to prior work on CIFAR-FS and miniImageNet.} 
    Average 5-way 5-shot classification accuracies on the test set with $95\%$ confidence intervals. We evaluate our method on a single run, where the accuracy is the mean accuracy of 250 randomly sampled tasks. Models with $^*$ use stronger data augmentation for the supervised objective. \textit{Sup.} refers to the supervised task and \textit{BYOL} to the representation prediction task in \citeauthor{grill2020bootstrap} (\citeyear{grill2020bootstrap}).}
    \centering
    \begin{tabular}{llcc}
    \toprule
    & & \multicolumn{2}{c}{Accuracy (\%)}\\
    \cmidrule{3-4}
    Method & Backbone & CIFAR-FS & miniImageNet \\
    \midrule
    Prototypical Networks \cite{snell2017prototypical} & ResNet-12 & $83.5 \pm 0.5$ & $78.63 \pm 0.48$\\
    TADAM \cite{oreshkin2018tadam} & ResNet-12 & $-$ & $76.70 \pm 0.30$\\
    MetaOptNet \cite{lee2019meta} & ResNet-12 & $84.3 \pm 0.5$ & $78.63 \pm 0.48$\\
    RFS-simple \cite{tian2020rethinking} & ResNet-12 & $86.0 \pm 0.5$ & $79.64 \pm 0.44$\\
    RFS-distill \cite{tian2020rethinking} & ResNet-12 & $\mathbf{86.9} \pm \mathbf{0.5}$ & $\mathbf{82.14} \pm \mathbf{0.43}$\\
    CC + rot \cite{gidaris2019boosting} & WRN-28-10 & $86.1 \pm 0.2$ & $79.87 \pm 0.33$\\
    \midrule
    Sup. & ResNet-12 & $87.65 \pm 0.75$ & $77.62 \pm 0.70$\\
    Sup.$^*$ + BYOL & ResNet-12 & $\mathbf{88.46} \pm \mathbf{0.68}$ & $\mathbf{78.30} \pm \mathbf{0.69}$\\
    \bottomrule
    \end{tabular}
    \label{tab:results-prior-work}
\end{table*}

\subsection{Avoiding Collapse with BYOL}
Although the related method BYOL reports poor performance when instantaneously updating the target network ($\tau = 0$) as it destabilizes training, we found that with an additional task to solve, the exponential moving average constant is not required to avoid representational collapse. On CIFAR-FS, we report an accuracy of $82.19 \pm 0.83\%$ for this run, compared to $82.51 \pm 0.82\%$ when using $\tau = 0.99$ on the same seed.

We hypothesize that solving an additional task in parallel with the BYOL loss (e.g. supervised objective) provides additional gradient that prevents collapse. Similar behaviour has been observed by \citeauthor{schwarzer2020dataefficient} (\citeyear{schwarzer2020dataefficient}) when using a reinforcement learning objective in parallel.

\section{Conclusion}

Based on the simple baseline of \citeauthor{tian2020rethinking} (\citeyear{tian2020rethinking}), we have proposed a multi-task framework with self-supervised auxiliary tasks to improve few-shot image classification. Based on our detailed experiments on CIFAR-FS and miniImageNet, we show that leveraging self-supervision improves transfer learning performance on novel classes. Our experiments show that the rotation prediction task, when paired with the supervised objective, improves few-shot performance, but that the representation prediction task (BYOL) is most beneficial. This result suggests that BYOL is a strong data-dependent regularizer by enforcing the representations for different views of the same image to be closer together in latent space. Furthermore, we show that these two tasks are mutually beneficial, and can be used in combination to improve few-shot classification performance under our multi-task framework. Finally, the proposed framework can be used efficiently when the transformations of the different tasks are shared to solve each task simultaneously.

\paragraph{Future work.} In this work, we sample images from the annotated set for the self-supervised tasks. On the other hand, these auxiliary tasks do not depend on class labels. Thus, one could easily extend to learn from additional unlabeled data and further improve few-shot performance. This makes our multi-task framework especially appealing in scenarios where labeled data is scarce. Future works could extend the use of our framework to the semi-supervised few-shot learning setting and leverage additional unlabeled data.


\bibliography{refs}

\begin{thebibliography}{54}
\providecommand{\natexlab}[1]{#1}
\providecommand{\url}[1]{\texttt{#1}}
\expandafter\ifx\csname urlstyle\endcsname\relax
  \providecommand{\doi}[1]{doi: #1}\else
  \providecommand{\doi}{doi: \begingroup \urlstyle{rm}\Url}\fi

\bibitem[Bachman et~al.(2019)Bachman, Hjelm, and
  Buchwalter]{bachman2019learning}
Bachman, P., Hjelm, R.~D., and Buchwalter, W.
\newblock Learning representations by maximizing mutual information across
  views.
\newblock In \emph{Advances in Neural Information Processing Systems}, pp.\
  15535--15545, 2019.

\bibitem[Bengio et~al.(1992)Bengio, Bengio, Cloutier, and
  Gecsei]{bengio1992slr}
Bengio, S., Bengio, Y., Cloutier, J., and Gecsei, J.
\newblock On the optimization of a synaptic learning rule.
\newblock In \emph{Optimality in Artificial and Biological Neural Networks},
  pp.\  6--8, 1992.

\bibitem[Bertinetto et~al.(2018)Bertinetto, Henriques, Torr, and
  Vedaldi]{bertinetto2018metalearning}
Bertinetto, L., Henriques, J.~F., Torr, P.~H., and Vedaldi, A.
\newblock Meta-learning with differentiable closed-form solvers.
\newblock \emph{arXiv preprint arXiv:1805.08136}, 2018.

\bibitem[Buitinck et~al.(2013)Buitinck, Louppe, Blondel, Pedregosa, Mueller,
  Grisel, Niculae, Prettenhofer, Gramfort, Grobler, Layton, VanderPlas, Joly,
  Holt, and Varoquaux]{sklearn_api}
Buitinck, L., Louppe, G., Blondel, M., Pedregosa, F., Mueller, A., Grisel, O.,
  Niculae, V., Prettenhofer, P., Gramfort, A., Grobler, J., Layton, R.,
  VanderPlas, J., Joly, A., Holt, B., and Varoquaux, G.
\newblock {API} design for machine learning software: experiences from the
  scikit-learn project.
\newblock In \emph{ECML PKDD Workshop: Languages for Data Mining and Machine
  Learning}, pp.\  108--122, 2013.

\bibitem[Chen et~al.(2020{\natexlab{a}})Chen, Kornblith, Norouzi, and
  Hinton]{chen2020simple}
Chen, T., Kornblith, S., Norouzi, M., and Hinton, G.
\newblock A simple framework for contrastive learning of visual
  representations.
\newblock In \emph{Proceedings of the 37th International Conference on Machine
  Learning}, pp.\  1597--1607, 2020{\natexlab{a}}.

\bibitem[Chen et~al.(2020{\natexlab{b}})Chen, Kornblith, Swersky, Norouzi, and
  Hinton]{chen2020big}
Chen, T., Kornblith, S., Swersky, K., Norouzi, M., and Hinton, G.~E.
\newblock Big self-supervised models are strong semi-supervised learners.
\newblock \emph{Advances in Neural Information Processing Systems}, 33,
  2020{\natexlab{b}}.

\bibitem[Chen et~al.(2020{\natexlab{c}})Chen, Fan, Girshick, and
  He]{chen2020improved}
Chen, X., Fan, H., Girshick, R., and He, K.
\newblock Improved baselines with momentum contrastive learning.
\newblock \emph{arXiv preprint arXiv:2003.04297}, 2020{\natexlab{c}}.

\bibitem[Cubuk et~al.(2019)Cubuk, Zoph, Mane, Vasudevan, and
  Le]{cubuk2019autoaugment}
Cubuk, E.~D., Zoph, B., Mane, D., Vasudevan, V., and Le, Q.~V.
\newblock Autoaugment: Learning augmentation strategies from data.
\newblock In \emph{Proceedings of the IEEE conference on computer vision and
  pattern recognition}, pp.\  113--123, 2019.

\bibitem[Cubuk et~al.(2020)Cubuk, Zoph, Shlens, and Le]{cubuk2020randaugment}
Cubuk, E.~D., Zoph, B., Shlens, J., and Le, Q.~V.
\newblock Randaugment: Practical automated data augmentation with a reduced
  search space.
\newblock In \emph{Proceedings of the IEEE/CVF Conference on Computer Vision
  and Pattern Recognition Workshops}, pp.\  702--703, 2020.

\bibitem[DeVries \& Taylor(2017)DeVries and Taylor]{devries2017improved}
DeVries, T. and Taylor, G.~W.
\newblock Improved regularization of convolutional neural networks with cutout.
\newblock \emph{arXiv preprint arXiv:1708.04552}, 2017.

\bibitem[Dhillon et~al.(2020)Dhillon, Chaudhari, Ravichandran, and
  Soatto]{dhillon2020baseline}
Dhillon, G.~S., Chaudhari, P., Ravichandran, A., and Soatto, S.
\newblock A baseline for few-shot image classification.
\newblock In \emph{Proceedings of the 8th International Conference on Learning
  Representations}, 2020.

\bibitem[Doersch \& Zisserman(2017)Doersch and Zisserman]{doersch2017multi}
Doersch, C. and Zisserman, A.
\newblock Multi-task self-supervised visual learning.
\newblock In \emph{Proceedings of the IEEE International Conference on Computer
  Vision}, pp.\  2051--2060, 2017.

\bibitem[Doersch et~al.(2015)Doersch, Gupta, and
  Efros]{doersch2015unsupervised}
Doersch, C., Gupta, A., and Efros, A.~A.
\newblock Unsupervised visual representation learning by context prediction.
\newblock In \emph{Proceedings of the IEEE international conference on computer
  vision}, pp.\  1422--1430, 2015.

\bibitem[Fetterman \& Albrecht(2020)Fetterman and
  Albrecht]{fetterman2020understanding}
Fetterman, A. and Albrecht, J.
\newblock Understanding self-supervised and contrastive learning with
  "bootstrap your own latent" (byol), 2020.
\newblock URL
  \url{https://www.untitled-ai.com/understanding-self-supervised-contrastive-learning.html}.

\bibitem[Finn et~al.(2017)Finn, Abbeel, and Levine]{finn2017model}
Finn, C., Abbeel, P., and Levine, S.
\newblock Model-agnostic meta-learning for fast adaptation of deep networks.
\newblock In \emph{Proceedings of the 34th International Conference on Machine
  Learning}, pp.\  1126--1135, 2017.

\bibitem[Ghiasi et~al.(2018)Ghiasi, Lin, and Le]{ghiasi2018dropblock}
Ghiasi, G., Lin, T.-Y., and Le, Q.~V.
\newblock Dropblock: A regularization method for convolutional networks.
\newblock \emph{Advances in neural information processing systems},
  31:\penalty0 10727--10737, 2018.

\bibitem[Gidaris \& Komodakis(2018)Gidaris and Komodakis]{gidaris2018dynamic}
Gidaris, S. and Komodakis, N.
\newblock Dynamic few-shot visual learning without forgetting.
\newblock In \emph{Proceedings of the IEEE Conference on Computer Vision and
  Pattern Recognition}, pp.\  4367--4375, 2018.

\bibitem[Gidaris et~al.(2018)Gidaris, Singh, and
  Komodakis]{gidaris2018unsupervised}
Gidaris, S., Singh, P., and Komodakis, N.
\newblock Unsupervised representation learning by predicting image rotations.
\newblock In \emph{Proceedings of the 6th International Conference on Learning
  Representations}, 2018.

\bibitem[Gidaris et~al.(2019)Gidaris, Bursuc, Komodakis, P{\'e}rez, and
  Cord]{gidaris2019boosting}
Gidaris, S., Bursuc, A., Komodakis, N., P{\'e}rez, P., and Cord, M.
\newblock Boosting few-shot visual learning with self-supervision.
\newblock In \emph{Proceedings of the IEEE International Conference on Computer
  Vision}, pp.\  8059--8068, 2019.

\bibitem[Grill et~al.(2020)Grill, Strub, Altch{\'e}, Tallec, Richemond,
  Buchatskaya, Doersch, Avila~Pires, Guo, Gheshlaghi~Azar,
  et~al.]{grill2020bootstrap}
Grill, J.-B., Strub, F., Altch{\'e}, F., Tallec, C., Richemond, P.,
  Buchatskaya, E., Doersch, C., Avila~Pires, B., Guo, Z., Gheshlaghi~Azar, M.,
  et~al.
\newblock Bootstrap your own latent-a new approach to self-supervised learning.
\newblock In \emph{Advances in Neural Information Processing Systems}, 2020.

\bibitem[He et~al.(2016)He, Zhang, Ren, and Sun]{he2016deep}
He, K., Zhang, X., Ren, S., and Sun, J.
\newblock Deep residual learning for image recognition.
\newblock In \emph{Proceedings of the IEEE conference on computer vision and
  pattern recognition}, pp.\  770--778, 2016.

\bibitem[He et~al.(2020)He, Fan, Wu, Xie, and Girshick]{he2020momentum}
He, K., Fan, H., Wu, Y., Xie, S., and Girshick, R.
\newblock Momentum contrast for unsupervised visual representation learning.
\newblock In \emph{Proceedings of the IEEE/CVF Conference on Computer Vision
  and Pattern Recognition}, pp.\  9729--9738, 2020.

\bibitem[Henaff(2020)]{henaff2020data}
Henaff, O.
\newblock Data-efficient image recognition with contrastive predictive coding.
\newblock In \emph{International Conference on Machine Learning}, pp.\
  4182--4192. PMLR, 2020.

\bibitem[Ioffe \& Szegedy(2015)Ioffe and Szegedy]{ioffe2015batch}
Ioffe, S. and Szegedy, C.
\newblock Batch normalization: Accelerating deep network training by reducing
  internal covariate shift.
\newblock In \emph{Proceedings of the 32nd International Conference on Machine
  Learning}, pp.\  448--456, 2015.

\bibitem[Larsson et~al.(2016)Larsson, Maire, and
  Shakhnarovich]{larsson2016learning}
Larsson, G., Maire, M., and Shakhnarovich, G.
\newblock Learning representations for automatic colorization.
\newblock In \emph{European conference on computer vision}, pp.\  577--593,
  2016.

\bibitem[Lee et~al.(2019)Lee, Maji, Ravichandran, and Soatto]{lee2019meta}
Lee, K., Maji, S., Ravichandran, A., and Soatto, S.
\newblock Meta-learning with differentiable convex optimization.
\newblock In \emph{Proceedings of the IEEE Conference on Computer Vision and
  Pattern Recognition}, pp.\  10657--10665, 2019.

\bibitem[Misra \& Maaten(2020)Misra and Maaten]{misra2020self}
Misra, I. and Maaten, L. v.~d.
\newblock Self-supervised learning of pretext-invariant representations.
\newblock In \emph{Proceedings of the IEEE/CVF Conference on Computer Vision
  and Pattern Recognition}, pp.\  6707--6717, 2020.

\bibitem[Nair \& Hinton(2010)Nair and Hinton]{nair2010rectified}
Nair, V. and Hinton, G.~E.
\newblock Rectified linear units improve restricted boltzmann machines.
\newblock In \emph{Proceedings of the 27th International Conference on
  International Conference on Machine Learning}, pp.\  807–814, 2010.

\bibitem[Noroozi \& Favaro(2016)Noroozi and Favaro]{noroozi2016unsupervised}
Noroozi, M. and Favaro, P.
\newblock Unsupervised learning of visual representations by solving jigsaw
  puzzles.
\newblock In \emph{European Conference on Computer Vision}, pp.\  69--84, 2016.

\bibitem[Oord et~al.(2018)Oord, Li, and Vinyals]{oord2018representation}
Oord, A. v.~d., Li, Y., and Vinyals, O.
\newblock Representation learning with contrastive predictive coding.
\newblock \emph{arXiv preprint arXiv:1807.03748}, 2018.

\bibitem[Oreshkin et~al.(2018)Oreshkin, L{\'o}pez, and
  Lacoste]{oreshkin2018tadam}
Oreshkin, B., L{\'o}pez, P.~R., and Lacoste, A.
\newblock Tadam: Task dependent adaptive metric for improved few-shot learning.
\newblock In \emph{Advances in Neural Information Processing Systems}, pp.\
  721--731, 2018.

\bibitem[Paszke et~al.(2019)Paszke, Gross, Massa, Lerer, Bradbury, Chanan,
  Killeen, Lin, Gimelshein, Antiga, Desmaison, Kopf, Yang, DeVito, Raison,
  Tejani, Chilamkurthy, Steiner, Fang, Bai, and Chintala]{pytorch2019neurips}
Paszke, A., Gross, S., Massa, F., Lerer, A., Bradbury, J., Chanan, G., Killeen,
  T., Lin, Z., Gimelshein, N., Antiga, L., Desmaison, A., Kopf, A., Yang, E.,
  DeVito, Z., Raison, M., Tejani, A., Chilamkurthy, S., Steiner, B., Fang, L.,
  Bai, J., and Chintala, S.
\newblock Pytorch: An imperative style, high-performance deep learning library.
\newblock In \emph{Advances in Neural Information Processing Systems 32}, pp.\
  8024--8035, 2019.

\bibitem[Qi et~al.(2018)Qi, Brown, and Lowe]{qi2018low}
Qi, H., Brown, M., and Lowe, D.~G.
\newblock Low-shot learning with imprinted weights.
\newblock In \emph{Proceedings of the IEEE conference on computer vision and
  pattern recognition}, pp.\  5822--5830, 2018.

\bibitem[Raghu et~al.(2020)Raghu, Raghu, Bengio, and Vinyals]{raghu2020rapid}
Raghu, A., Raghu, M., Bengio, S., and Vinyals, O.
\newblock Rapid learning or feature reuse? towards understanding the
  effectiveness of maml.
\newblock In \emph{Proceedings of the 8th International Conference on Learning
  Representations}, 2020.

\bibitem[Ravi \& Larochelle(2017)Ravi and Larochelle]{ravi2016optimization}
Ravi, S. and Larochelle, H.
\newblock Optimization as a model for few-shot learning.
\newblock In \emph{Proceedings of the 5th International Conference on Learning
  Representations}, 2017.

\bibitem[Riba et~al.(2020)Riba, Mishkin, Ponsa, Rublee, and
  Bradski]{riba2020kornia}
Riba, E., Mishkin, D., Ponsa, D., Rublee, E., and Bradski, G.
\newblock Kornia: an open source differentiable computer vision library for
  pytorch.
\newblock In \emph{The IEEE Winter Conference on Applications of Computer
  Vision}, pp.\  3674--3683, 2020.

\bibitem[Rusu et~al.(2019)Rusu, Rao, Sygnowski, Vinyals, Pascanu, Osindero, and
  Hadsell]{rusu2019meta}
Rusu, A.~A., Rao, D., Sygnowski, J., Vinyals, O., Pascanu, R., Osindero, S.,
  and Hadsell, R.
\newblock Meta-learning with latent embedding optimization.
\newblock In \emph{Proceedings of the 7th International Conference on Learning
  Representations}, 2019.

\bibitem[Schmidhuber(1987)]{schmidhuber1987srl}
Schmidhuber, J.
\newblock Evolutionary principles in self-referential learning. on learning now
  to learn: The meta-meta-meta...-hook.
\newblock Diploma thesis, Technische Universitat Munchen, Germany, 14 May 1987.

\bibitem[Schwarzer et~al.(2020)Schwarzer, Anand, Goel, Hjelm, Courville, and
  Bachman]{schwarzer2020dataefficient}
Schwarzer, M., Anand, A., Goel, R., Hjelm, R.~D., Courville, A., and Bachman,
  P.
\newblock Data-efficient reinforcement learning with self-predictive
  representations.
\newblock \emph{arXiv preprint arXiv:2007.05929}, 2020.

\bibitem[Snell et~al.(2017)Snell, Swersky, and Zemel]{snell2017prototypical}
Snell, J., Swersky, K., and Zemel, R.
\newblock Prototypical networks for few-shot learning.
\newblock In \emph{Advances in neural information processing systems}, pp.\
  4077--4087, 2017.

\bibitem[Srivastava et~al.(2014)Srivastava, Hinton, Krizhevsky, Sutskever, and
  Salakhutdinov]{srivastava2014dropout}
Srivastava, N., Hinton, G., Krizhevsky, A., Sutskever, I., and Salakhutdinov,
  R.
\newblock Dropout: A simple way to prevent neural networks from overfitting.
\newblock \emph{Journal of Machine Learning Research}, 15\penalty0
  (56):\penalty0 1929--1958, 2014.

\bibitem[Su et~al.(2020)Su, Maji, and Hariharan]{su2020does}
Su, J.-C., Maji, S., and Hariharan, B.
\newblock When does self-supervision improve few-shot learning?
\newblock In \emph{European Conference on Computer Vision}, pp.\  645--666.
  Springer, 2020.

\bibitem[Sung et~al.(2018)Sung, Yang, Zhang, Xiang, Torr, and
  Hospedales]{sung2018learning}
Sung, F., Yang, Y., Zhang, L., Xiang, T., Torr, P.~H., and Hospedales, T.~M.
\newblock Learning to compare: Relation network for few-shot learning.
\newblock In \emph{Proceedings of the IEEE Conference on Computer Vision and
  Pattern Recognition}, pp.\  1199--1208, 2018.

\bibitem[Thrun \& Pratt(1998)Thrun and Pratt]{thrun1998ltl}
Thrun, S. and Pratt, L.
\newblock \emph{Learning to Learn}.
\newblock Springer Science \& Business Media, 1998.

\bibitem[Tian et~al.(2019)Tian, Krishnan, and Isola]{tian2019contrastive}
Tian, Y., Krishnan, D., and Isola, P.
\newblock Contrastive multiview coding.
\newblock \emph{arXiv preprint arXiv:1906.05849}, 2019.

\bibitem[Tian et~al.(2020{\natexlab{a}})Tian, Wang, Krishnan, Tenenbaum, and
  Isola]{tian2020rethinking}
Tian, Y., Wang, Y., Krishnan, D., Tenenbaum, J.~B., and Isola, P.
\newblock Rethinking few-shot image classification: a good embedding is all you
  need?
\newblock \emph{arXiv preprint arXiv:2003.11539}, 2020{\natexlab{a}}.

\bibitem[Tian et~al.(2020{\natexlab{b}})Tian, Yu, Chen, and
  Ganguli]{tian2020understanding}
Tian, Y., Yu, L., Chen, X., and Ganguli, S.
\newblock Understanding self-supervised learning with dual deep networks.
\newblock \emph{arXiv preprint arXiv:2010.00578}, 2020{\natexlab{b}}.

\bibitem[van~der Maaten \& Hinton(2008)van~der Maaten and
  Hinton]{vandermaaten2008vis}
van~der Maaten, L. and Hinton, G.
\newblock Visualizing data using t-sne.
\newblock \emph{Journal of Machine Learning Research}, 9\penalty0
  (86):\penalty0 2579--2605, 2008.

\bibitem[Vinyals et~al.(2016)Vinyals, Blundell, Lillicrap, Wierstra,
  et~al.]{vinyals2016matching}
Vinyals, O., Blundell, C., Lillicrap, T., Wierstra, D., et~al.
\newblock Matching networks for one shot learning.
\newblock In \emph{Advances in neural information processing systems}, pp.\
  3630--3638, 2016.

\bibitem[Wu et~al.(2018)Wu, Xiong, Yu, and Lin]{wu2018unsupervised}
Wu, Z., Xiong, Y., Yu, S.~X., and Lin, D.
\newblock Unsupervised feature learning via non-parametric instance
  discrimination.
\newblock In \emph{Proceedings of the IEEE Conference on Computer Vision and
  Pattern Recognition}, pp.\  3733--3742, 2018.

\bibitem[Yamaguchi et~al.(2019)Yamaguchi, Kanai, Shioda, and
  Takeda]{yamaguchi2019multiple}
Yamaguchi, S., Kanai, S., Shioda, T., and Takeda, S.
\newblock Multiple pretext-task for self-supervised learning via mixing
  multiple image transformations.
\newblock \emph{arXiv preprint arXiv:1912.11603}, 2019.

\bibitem[Yun et~al.(2019)Yun, Han, Oh, Chun, Choe, and Yoo]{yun2019cutmix}
Yun, S., Han, D., Oh, S.~J., Chun, S., Choe, J., and Yoo, Y.
\newblock Cutmix: Regularization strategy to train strong classifiers with
  localizable features.
\newblock In \emph{Proceedings of the IEEE International Conference on Computer
  Vision}, pp.\  6023--6032, 2019.

\bibitem[Zhang et~al.(2018)Zhang, Cisse, Dauphin, and
  Lopez-Paz]{zhang2018mixup}
Zhang, H., Cisse, M., Dauphin, Y.~N., and Lopez-Paz, D.
\newblock mixup: Beyond empirical risk minimization.
\newblock In \emph{Proceedings of the 6th International Conference on Learning
  Representations}, 2018.

\bibitem[Zhang et~al.(2016)Zhang, Isola, and Efros]{zhang2016colorful}
Zhang, R., Isola, P., and Efros, A.~A.
\newblock Colorful image colorization.
\newblock In \emph{European conference on computer vision}, pp.\  649--666,
  2016.

\end{thebibliography}
\bibliographystyle{icml2020}

\clearpage
\section*{Appendix}
\appendix

\section{Regularizing for Stability} \label{appendix:exp-no-aug}
\begin{figure}[th!]
    \centering
    \includegraphics[width=0.45\textwidth]{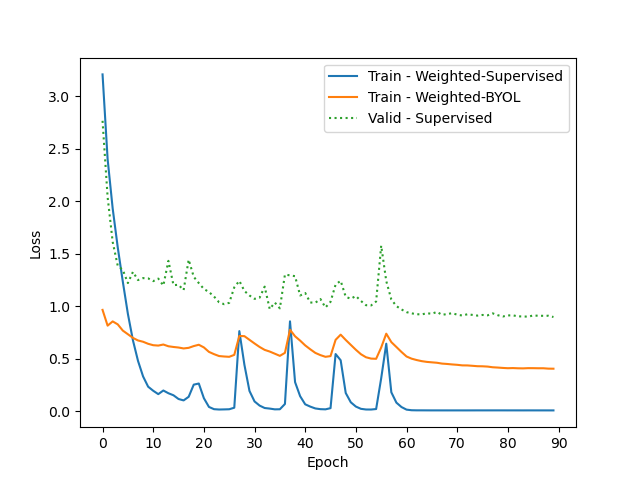}
    \caption{Learning curves of our multi-task framework, using a supervised objective (no data augmentation) and BYOL auxiliary task.
    Losses of each task during the training of the BYOL task and supervised task without any data augmentation over 90 epochs.
    Training seems to stabilize after epoch 60, where the learning rate was decayed.}
    \label{fig:byol-unstable}
\end{figure}

During our initial experiments, we tried combining the supervised task without any data augmentation with the BYOL auxiliary task. The goal of this experiment was to find out if BYOL was sufficient to enforce invariance to the transformations used as part of its data augmentation strategy, without directly applying any transformations to the inputs of the supervised task. As shown in Figure \ref{fig:byol-unstable} the training was highly unstable, with the two tasks competing against each other. This resulted in a performance worse than the supervised baseline, achieving a few-shot accuracy of $78.98 \pm 0.92\%$ with a ResNet-18 backbone. We believe additional regularization (e.g. Dropout \cite{srivastava2014dropout}) to the backbone could have mitigated this effect, but this is still an interesting finding.

\section{Qualitative Analysis}

\begin{figure}[th!]
    \centering
    \includegraphics[width=0.45\textwidth]{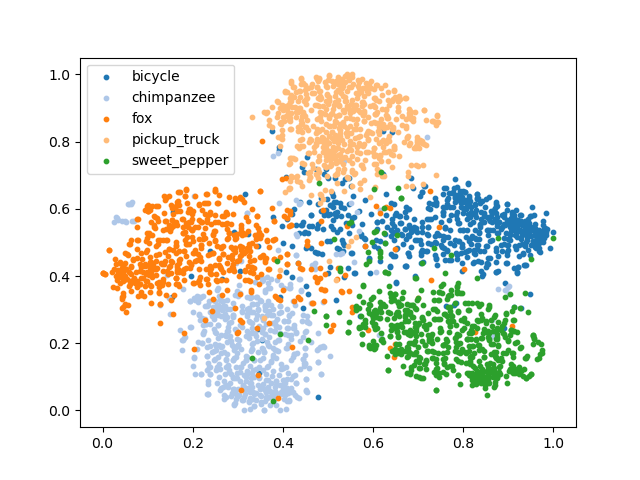}
    \caption{\textbf{t-SNE visualization of the embedding space for a sampled 5-way task (unseen) on CIFAR-FS.} The embeddings result from a model trained with the supervised and BYOL tasks.
}
    \label{fig:annex:t-sne}
\end{figure}

We are interested in learning a linearly separable representation of the input so that a simple linear classifier can be trained on unseen classes and discriminate between them. Hence, it is possible to proceed to a qualitative evaluation of the learned representations by visualizing the embedding space. As illustrated by the t-SNE \cite{vandermaaten2008vis} visualization in Figure \ref{fig:annex:t-sne}, multiple different clusters are easily identifiable even though there are some points that are still intertwined with others points that belong to other classes. The fact that such clusters exist on unseen classes is quite interesting, and reinforces the statement that a good representation of the data is extremely important in few-shot learning \cite{raghu2020rapid, tian2020rethinking}.

\section{Additional Supervised + BYOL Experiment} \label{appendix:extra-sup-byol}
In Table \ref{tab:extra-sup-byol}, we show the difference in performance when leveraging all (2) augmented views generated for the BYOL task with the supervised objective as well. Effectively, the supervised loss is computed on 2 augmented views for each image in the batch instead of only the first augmented view. Results were inconclusive.
\begin{table}[th!]
    \caption{\textbf{Difference in performance when leveraging all augmented views for the supervised objective when used in combination with BYOL.}
    Average 5-way 5-shot classification accuracies on the test set with $95\%$ confidence intervals. The backbone used for these experiments is ResNet-18.
    }
    \centering
    \begin{tabular}{lcc}
    \toprule
    Dataset & Sup. per image & Accuracy (\%) \\
    \midrule
    CIFAR-FS      & 1 & $\mathbf{82.43} \pm \mathbf{0.92}$\\
    CIFAR-FS      & 2 & $82.33 \pm 0.89$\\
    \midrule
    miniImageNet & 1 & $71.47 \pm 0.90$\\
    miniImageNet & 2 & $\mathbf{71.95} \pm \mathbf{0.78}$\\
    \bottomrule
    \end{tabular}
    \label{tab:extra-sup-byol}
\end{table}

\section{Image Augmentations} \label{appendix:data-aug}
The image augmentations parameters for the different settings are listed in Listing \ref{lst:data-aug-code}. Default augmentations are the same as \cite{tian2020rethinking}, and hard augmentations are for the BYOL task (and supervised task in some experiments).

    
\begin{listing*}[ht!]
\caption{Data augmentation code with parameters.} 
\begin{minted}{python}
import random
import kornia.augmentation as K
import kornia.filters as F

class RandomApply(object):
    def __init__(self, fn, p):
        self.fn = fn
        self.p = p

    def __call__(self, x):
        if random.random() > self.p:
            return x
        return self.fn(x)

size = (96, 96) if CIFAR_FS else (84, 84)
pad = 4 if CIFAR_FS else 8

default_tfm = [
    K.RandomCrop(size, padding=pad),
    K.ColorJitter(brightness=0.4, contrast=0.4, saturation=0.4, p=1.0),
    K.RandomHorizontalFlip(p=0.5)
]

hard_tfm = [
    K.ColorJitter(brightness=0.4, contrast=0.4, saturation=0.4, hue=0.1, p=0.8),
    K.RandomGrayscale(p=0.2),
    K.RandomHorizontalFlip(p=0.5),
    RandomApply(F.GaussianBlur2d(kernel_size=(3, 3), sigma=(1.5, 1.5)), p=0.1),
    K.RandomResizedCrop(size, scale=(0.35, 1.0), p=1.0) 
]
\end{minted}
\label{lst:data-aug-code}
\end{listing*}

\end{document}